\documentclass[11pt]{article}
\usepackage[round]{natbib}

\usepackage{amssymb,amsmath, amsfonts}
\usepackage{graphicx}
\usepackage{bm,color}
\usepackage{algorithm, algpseudocode}
\usepackage{lscape}
\usepackage{cite}
\usepackage[unicode,final]{hyperref}
\usepackage{fancyhdr}

\allowdisplaybreaks  

\newcommand{\argmin}{\mathop{\rm argmin}\limits}
\newcommand{\argmax}{\mathop{\rm argmax}\limits}

\providecommand{\doi}[1]{DOI~\href{http://dx.doi.org/#1}{\path|#1|}}

\begin{document}

{\begin{flushleft}
{\LARGE Cost-effective Framework for Gradual Domain Adaptation with Multifidelity}
\end{flushleft}
\ \\
{\bf \large 
Shogo Sagawa$^{1}$ and
Hideitsu Hino$^{2,3}$ 
}

\begin{flushleft}
$^{1}$ Department of Statistical Science,
School of Multidisciplinary Sciences, The Graduate University for Advanced Studies (SOKENDAI)\\
Shonan Village, Hayama, Kanagawa 240-0193, Japan\\
$^{2}$Department of Statistical Modeling, The Institute of Statistical Mathematics\\
10-3 Midori-cho, Tachikawa, Tokyo 190-8562, Japan\\
$^{3}$Center for Advanced Intelligence Project(AIP), RIKEN,\\
1-4-4 Nihonbashi, Chuo-ku, Tokyo 103-0027, Japan
\end{flushleft}}
\thispagestyle{fancy}
\rhead{}
\lhead{}

\begin{center} {\bf Abstract} \end{center}
In domain adaptation, when there is a large distance between the source and target domains, the prediction performance will degrade. Gradual domain adaptation is one of the solutions to such an issue, assuming that we have access to intermediate domains, which shift gradually from the source to the target domain. In previous works, it was assumed that the number of samples in the intermediate domains was sufficiently large; hence, self-training was possible without the need for labeled data. If the number of accessible intermediate domains is restricted, the distances between domains become large, and self-training will fail. Practically, the cost of samples in intermediate domains will vary, and it is natural to consider that the closer an intermediate domain is to the target domain, the higher the cost of obtaining samples from the intermediate domain is. To solve the trade-off between cost and accuracy, we propose a framework that combines multifidelity and active domain adaptation. The effectiveness of the proposed method is evaluated by experiments with real-world datasets.

\section{Introduction}\label{sec:intro}
In the standard prediction problems using machine learning models, it is assumed that the test data in operation are samples obtained from the same probability distribution as the training data. It is known that the prediction performance deteriorates when the assumption is broken. The simplest solution is to discard the training data and construct a new machine learning model by collecting new samples obtained from the same probability distribution as the prediction target. However, obtaining enough samples to construct a new machine learning model generally requires a large amount of cost.

Transfer learning~\citep{zhuang2020comprehensive} is a technique to effectively use of existing data, similarly to human thinking that makes use of past experiences. In transfer learning, when the probability distributions of the training data and the prediction target are different but the task of machine learning is the same, the problem is known as domain adaptation~\citep{ben2007analysis}. In domain adaptation, the training data distribution with rich labeled data is called the source domain, and the distribution of the prediction target is called the target domain. The case where a small number of labels are available from the target domain is called semi-supervised domain adaptation~\citep{wang2018deep}, and the case where there are no labels at all is called unsupervised domain adaptation~\citep{wilson2020survey}. The problem of unsupervised domain adaptation is more difficult and has been the subject of much research, including theoretical analysis~\citep{mansour2009domain, cortes2010learning, redko2020survey, zhao2019learning}.

In domain adaptation, the distance between domains significantly affects the predictive performance for the target data. Several distance metrics have been proposed, in this paper, following the previous work~\citep{kumar2020understanding}, we consider maximum per-class Wasserstein-infinity distance~\citep{villani2009optimal}. \citet{kumar2020understanding} assumed that even when the distance between the source and the target domains is large, the shift from the source domain to the target domain occurs gradually, and they proposed gradual domain adaptation. Gradual domain adaptation has the potential to make it possible to predict data in the target domain that is far different from the source domain by adding data from intermediate domains that interpolate between the source and target pairs.

The self-training for gradual domain adaptation has limited applicability because the distance between intermediate domains can be large in practical situations. We propose the application of active domain adaptation, which combines domain adaptation with active learning, as a solution to this problem. In reality, each query from each domain comes with some cost, and the cost is not uniform. In domain adaptation, we consider the case that the reason why the target dataset has no label is because of its high cost. Moreover, it is natural to assume that the cost of obtaining data from the intermediate domain is higher when it is closer to the target domain and lower when it is closer to the source domain. Note that the intermediate domains are given, and we cannot control the degree of domain shift. 

\begin{figure}[!htbp]
\centering
  \includegraphics[clip, width=11cm]{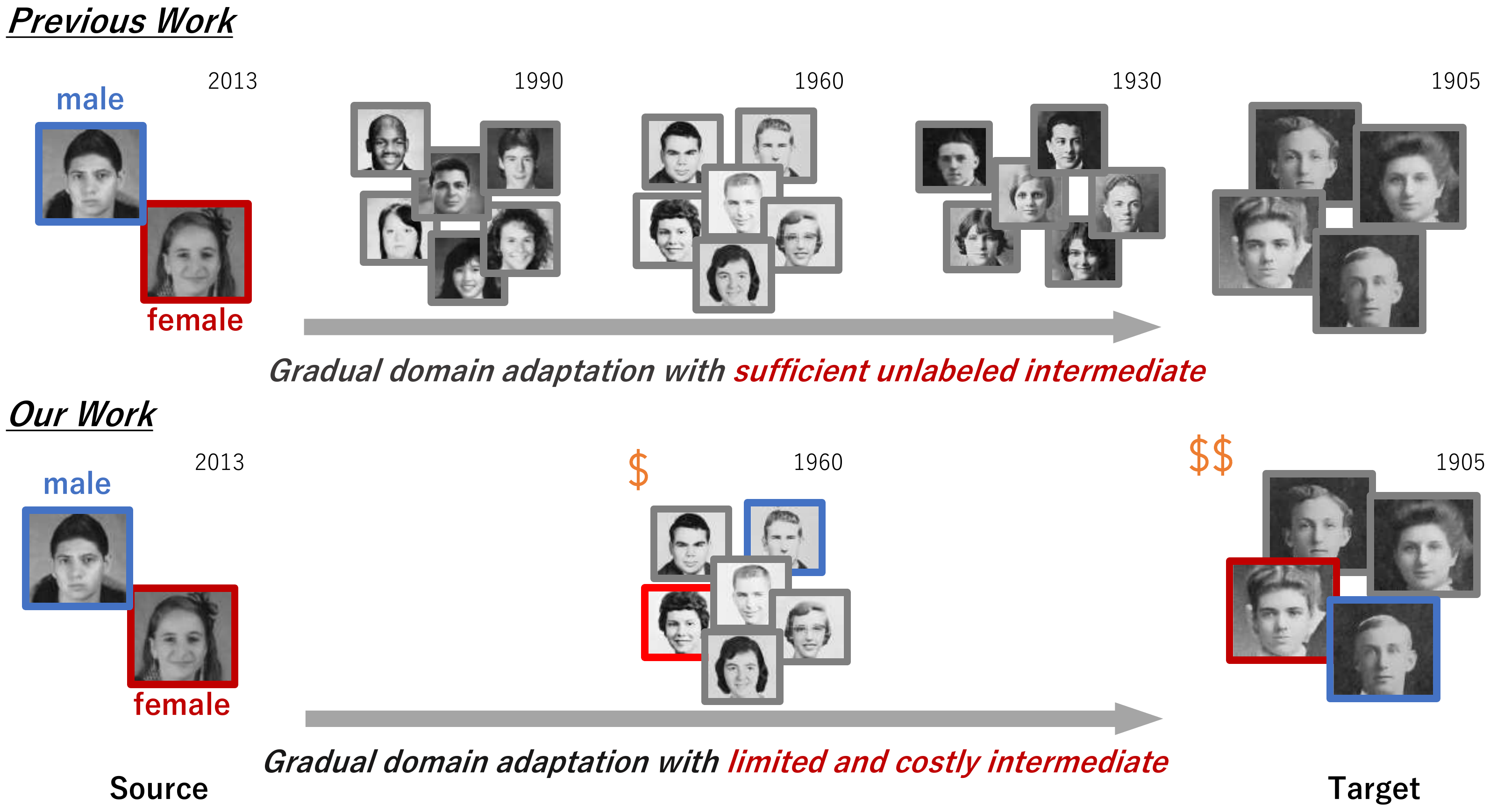}
  \caption{Our problem settings and proposed method with an example of the classification between male and female using portraits. To tackle the gradual domain shift in the situation where access to the intermediate domains is restricted and self-training is not possible, we request a small number of queries with different cost for each domain. A gray frame around the photo indicates that it is unlabeled, while a red or blue one indicates that it is labeled.}\label{fig:scheme}
\end{figure}

We show an example of our considered problem with the dataset of portraits in Figure~\ref{fig:scheme}.
Portraits is a dataset of images from yearbooks of American high school students from 1905 to 2013, and the task is to estimate the gender from the images. We consider images from 2013 as the source and images from 1905 as the target domain, respectively. Naturally, we can consider the images from each year as an intermediate domain since the images come from the yearbook. Assuming that an annotator gave us labels for the source domain, we can ask the annotator to annotate the images from the intermediate and the target domain. The annotator determines the gender of the person by confirming the record of the individual or by contact with the individual. It is natural to assume that the annotation of the older images is more difficult since the person cannot be contacted or the records of the individual may have been lost. Hence, the query cost for the older images which will be similar to images in the target domain is high, and the query cost of the newer images is low. Although we can obtain a reliable classifier by training with sufficient labeled data from the target domain, it is not cost-effective.

To resolve this trade-off between cost and uncertainty, by utilizing the notion of multifidelity learning~\citep{peherstorfer2018survey}, we propose gradual domain adaptation with multifidelity learning (GDAMF) as an active domain adaptation framework. The features of our proposed method are as follows.
\begin{enumerate}
   \item Since it is a gradual domain adaptation method, domain adaptation is possible even when the distance between the source and target domains is large.
   \item When the distances between domains are large, the performance of the previous method~\citep{kumar2020understanding} deteriorates. Our proposed method tackles this problem by requesting a small number of efficient queries from each domain.
    \item We considered a situation where queries from each domain come with some cost. The proposed method is cost-effective since it utilizes multiflidelity active learning.
    \item Since the proposed method utilizes the source, intermediate, and target data, it achieves a higher prediction performance than a method that uses a model trained only with the target data.
\end{enumerate}

The rest of the paper is organized as follows. We review some related works in Section~\ref{sec:rw}, then introduce the previous work as gradual domain adaptation in Section~\ref{sec:previous}. We propose our method in Section~\ref{sec:prop}. In Section~\ref{sec:exp}, we present the experiments. We describe the applicability of our proposed method and conclusion in Section~\ref{sec:disc&conc}.

\section{Related works}\label{sec:rw}
\subsection{Domain adaptation}\label{sec:rw-da}
In the context of domain adaptation, several methods have been developed to access one intermediate domain to improve the predictive power on the target dataset~\citep{hsu2020progressive, choi2020visual, cui2020gradually, dai2021idm, gong2019dlow, gadermayr2018gradual}.
In these methods, generative models are used such as generative adversarial networks (GAN)~\citep{pan2019recent} and normalizing flow~\citep{papamakarios2021normalizing} and the datasets from the source and target domains are mixed at arbitrary ratios to obtain the intermediate dataset. This procedure is regarded as learning common features shared between both the source and target domains.

The multi-source domain adaptation~\citep{zhao2020multi} and gradual domain adaptation are similar, though, the assumed situations are different. In multi-source domain adaptation, it is assumed that labeled data are collected from multiple different distributions. One simple approach to utilizing multiple source domains is combining the source domains into a single source domain. Since this simple approach does not provide good prediction performance, various approaches have been studied. One of the standard approaches of the multi-source domain adaptation assumes that the target distribution can be approximated by a mixture of multiple source distributions. Therefore, a combination of classifiers with weight is used for multi-source domain adaptation.
In gradual domain adaptation, it is assumed that we can utilize intermediate domains, which shift gradually from the source to the target domain. Even if the source and the target domain are not highly related (i.e., there is a large gap between the source and the target domain), we can transfer the learned knowledge on the source domain from the source to the target domain through the intermediate domains. In practice, we cannot adjust the degree of shift and cannot select which intermediate domain to use, and have no labels on intermediate domains. However, the ordered sequence of intermediate domains is given.
\citet{pmlr-v119-wang20h} proposed a method of multi-source domain adaptation with sequence. Their assumed situation corresponds to the situation in gradual domain adaptation, where all the labels of intermediate domains are given. The continuous change from the source to the target domain is learned under the condition that the source domain is finely partitioned and indices for identifying the intermediate domains are given.

\citet{kumar2020understanding} demonstrated that self-training domain adaptation is possible under the condition that there are intermediate domains whose sequence is given. In addition, theoretical analysis is also performed to show the upper bounds of self-training. An improved generalization error bound under more general assumptions was derived by ~\citep{pmlr-v162-wang22n}. \citet{dong2022algorithms} also conducted a theoretical analysis under a supervised gradual domain adaptation setting.

The situation where the intermediate domain exists but the index is unknown is considered in~\citep{zhang2021gradual,chen2021gradual}, and a method to align the intermediate domain is proposed.
In \citep{zhang2021gradual}, a method to artificially generate datasets with gradual domain change is proposed. In the method, a subset of data with high prediction reliability is sampled from the source dataset and the target dataset in predetermined proportions, and the mixed subsamples are used as an intermediate dataset. Note that this dataset is labeled by the ground-truth label for source-originated data and the predicted label for the target-originated data. The number of updates of the model is given as a hyperparameter, and the percentage of the target dataset is increased as the number of updates increases to represent the gradual change.
\citet{chen2021gradual} proposed a fine-tuning method to minimize the cycle consistency of self-training by giving a coarse domain discriminator as the initial guess of the domain index for each datum. It is shown that this method can cope with the case where the index of the intermediate domain is unknown or where data unsuitable for the intermediate domain are included. However, it is also noted that the prediction accuracy decreases significantly when the interval between intermediate domains is large. 
In~\citep{abnar2021gradual}, the authors consider the situation where there is a large distance between the source and target domains and the intermediate domain is not directly accessible. They propose to mix the source and target data at an arbitrary ratio, and use as the data from the intermediate domain. A possibility of self-training is also mentioned.

Several methods have been developed as gradual semi-supervised domain adaptation methods that utilize self-training and requests of queries~\citep{9681347,zhou2022online}. \citet{9681347} proposed a gradual semi-supervised domain adaptation method and provided a new dataset suitable for gradual domain adaptation. In online recommendation models, \citet{pmlr-v180-ye22b} proposed a method for temporal domain generalization, where the labels of the datasets corresponding to the intermediate domains are given.

Active domain adaptation, which combines domain adaptation with active learning, was proposed in~\citep{rai2010domain, saha2011active}. In active domain adaptation, the aim is to obtain a higher predictive power than in the case of using domain adaptation alone by querying the target dataset taking its predictive uncertainty into account.
Active domain adaptation has since been extended to deep learning~\citep{zhou2021discriminative, su2020active}. The querying strategies range from those based on clustering~\citep{prabhu2021active} to margins of classifiers. Batch query methods are also proposed in ~\citep{Plitz2015DistanceBA,de2021discrepancy}. 

\subsection{Multifidelity learning}\label{sec:rw-mf}
Multifidelity learning is a technique that has been developed mainly in the field of simulation to alleviate the problem of high computational cost for obtaining highly accurate simulation results, and its combination to active learning and Bayesian optimization are also considered~\citep{DBLP:journals/corr/abs-2012-00901,takeno2020multi}. Specifically, by combining simulators with various accuracies and costs, it aims to achieve the comparable accuracy and lower costs as those of using only high-accuracy simulators. It is also gaining importance in the relation to the diversity of labelers problem in crowdsourcing~\citep{huang2017cost}. A survey of multifidelity learning can be found in~\citep{peherstorfer2018survey}. To construct a model that integrates multiple fidelities in multifidelity learning, studies to improve the representation of correlations between fidelities~\citep{li2020multi} and have deal with heterogeneous inputs~\citep{hebbal2021multi, sarkarmulti} have been conducted. There have also been studies on high-dimensional outputs to handle large-scale simulations~\citep{wang2021multi,penwarden2021multifidelity}. Examples of combining multifidelity learning with Bayesian optimization and active learning have been described in several papers~\citep{grassi2021resource, dhulipala2021active, tran2020smf}, but the acquisition function used does not take into account the correlation and cost between simulators. The acquisition function used is one that determines the sample to be queried by considering the correlation between simulators and cost. Our proposed GDAMF differs in that it uses an acquisition function designed to improve the accuracy of classifications as well as the usual domain adaptation.

In this paper, we consider gradual active domain adaptation, which consists of sequential domains with different query costs. To the best of authors' knowledge, there has been no study on active domain adaptation that queries from sequential domains considering the cost. We note that since the conventional active domain adaptation methods can also be implemented as gradual domain adaptation by sequentially executing them, we conduct experiments to compare the results with those in~\citep{rai2010domain} in Section~\ref{sec:exp}.
\begingroup
\renewcommand{\arraystretch}{1.2} 
\begin{table*}[!htbp]
\caption{Notations}\label{tab:notations}
    \centering
    \scalebox{1.0}{
    \begin{tabular}{|l|p{9cm}|}  \hline
    $X \in \mathcal{X} = \mathbb{R}^{d}$ & random variables of the feature \\ \hline
    $\bm{x}$ & realizations of random variable $X$ \\ \hline
    $Y \in \mathcal{Y} = \{1, 2, \ldots, L\}$ & random variables of the label\\ \hline
    $y$ & realizations of random variable $Y$ \\ \hline
    $i, \; 1 \leq i \leq n_j$ & $i$-th observed sample with labels \\ \hline
    $i', \; 1 \leq i' \leq n'_j$ & $i'$-th observed sample with no labels \\ \hline
    $(j), \; 0 \leq j \leq K$ & $j$-th domain \\ \hline
    $D^{(j)}=\{(\bm{x}_i^{(j)}, y_i^{(j)})\}_{i=1}^{n_j}$ & labeled dataset of $j$-th domain \\ \hline
    $U^{(j)}=\{\bm{x}_{i'}^{(j)}\}_{i'=1}^{n'_{j}}$ & unlabeled dataset of $j$-th domain \\ \hline
    $\theta \in \Theta$ & parameter of classifier \\ \hline
    $h_{\theta^{(j)}}$ & classifier for $j$-th domain \\ \hline
    $c^{(j)} \in \mathbb{R}$ & query cost of $j$-th domain \\ \hline
    $m^{(j)} \in \mathbb{N}$ & number of query samples from $j$-th domain \\ \hline
    $r^{(j)}$ & ratio of the number of query samples from the target domain to the number of query samples from the $j$-th intermediate domain \\ \hline
    \end{tabular}}
\end{table*}
\endgroup

\section{Gradual domain adaptation}\label{sec:previous}
First, we introduce the gradual domain adaptation algorithm proposed by \citet{kumar2020understanding} and modify the algorithm in Section~\ref{sec:prop}. The notations of the symbols used in this paper are summarized in Table~\ref{tab:notations}. 

We consider a multi-class classification problem where $X \in \mathcal{X} = \mathbb{R}^{d}$ and $Y \in \mathcal{Y} = \{1, 2, \ldots, L\}$ be random variables of the feature and label, respectively. The realizations of each random variable are $\bm{x}$ and $y$, respectively. Let $D^{(j)}=\{(\bm{x}_i^{(j)}, y_i^{(j)})\}_{i=1}^{n_j}$ and $U^{(j)}=\{\bm{x}_{i'}^{(j)}\}_{i'=1}^{n'_{j}}$ be labeled and unlabeled datasets, respectively. The subscript $i, \; 1 \leq i \leq n_j$ means the $i$-th observed sample with labels, and the subscript $i', \; 1 \leq i' \leq n'_j$ means the $i'$-th observed sample with no labels, and the superscript $(j), \; 0 \leq j \leq K$ means the $j$-th domain. We assign the source and the target domain to $j=0$ and $j=K$, respectively. Note that the intermediate domains and the target domain have no labeled dataset and the source domain has no unlabeled dataset. We are considering the situation that the ordered sequence of the intermediate domains is given; hence, the domain assigned a small $j$ is close to the source domain. On the contrary, when $j$ is large, the domain is close to the target domain. In the previous work~\citep{kumar2020understanding}, the distance between two domains is defined as the maximum per-class Wasserstein-infinity distance~\citep{villani2009optimal}, and we also use this metric.

Let $h_{\theta}$ be a classifier with a parameter $\theta \in \Theta$, and the classifier outputs a conditional probability
\begin{align*}
    h_{\theta}(\bm{x}) = \begin{bmatrix}
                         \Pr(y=1|\bm{x},\theta)\\
                         \vdots \\
                         \Pr(y=L|\bm{x},\theta)
                         \end{bmatrix}
    \; , \; \sum_{l=1}^{L} \Pr(y=l|\bm{x},\theta) = 1.
\end{align*}
Denote by $\ell: \mathcal{Y} \times \mathcal{Y} \to \mathbb{R}_{\geq 0}$ the loss function. We obtain a parameter of the base model $\theta^{(0)}$ by minimizing the loss on the source dataset
\begin {equation}
    \theta^{(0)} = \argmin_{\theta \in \Theta} \frac{1}{n_{0}} \sum_{i=1}^{n_{0}} \ell(h_{\theta}(\bm{x}_i^{(0)}), y_i^{(0)})
    \label{eq:base-train}
\end {equation}
Similar to standard self-training, we use a pseudo-label for model updating. We define the sharpen function, which returns the pseudo-label by inputting the output of a model.
\begin {equation}
    {\mathrm{sharpen}}(h_{\theta}(\bm{x})) = \argmax_{l \in \{1,\ldots,L\}} \Pr(y=l|\bm{x},\theta)
\end {equation}
Given a parameter of the current model $\theta^{(j)}$ and unlabeled dataset in the next domain $U^{(j+1)}$, $\mathrm{ST}(\theta^{(j)}, U^{(j+1)})$ denotes the output of self-training.
\begin{equation}
    \mathrm{ST}(\theta^{(j)}, U^{(j+1)}) = \argmin_{\theta \in \Theta} \frac{1}{n'_{j+1}} \sum_{i'=1}^{n'_{j+1}} \ell(h_{\theta}(\bm{x}_{i'}^{(j+1)}), \mathrm{sharpen}(h_{\theta^{(j)}}(\bm{x}_{i'}^{(j+1)})))
    \label{eq:self-train}
\end{equation}
Since the sequence of the intermediate domains is given, we can apply self-training to the base model with the sequence of unlabeled datasets $U^{(1)}, U^{(2)}, \ldots, U^{(K)}$. Finally, we obtain the model parameter for the target domain $\theta^{(K)}$ 

\begin{align*}
    \theta^{(1)} &= \mathrm{ST}(\theta^{(0)}, U^{(1)}) \\
    \theta^{(2)} &= \mathrm{ST}(\theta^{(1)}, U^{(2)}) \\
    \vdots \\
    \theta^{(K)} &= \mathrm{ST}(\theta^{(K-1)}, U^{(K)}).
\end{align*}
Since we consider the problem where the number of accessible intermediate domains is limited, the distances between the domains are large. This problem will restrict the applicability of self-training, and we show a failed example of self-training under the condition that the distances between the domains are large in Section~\ref{sec:exp}. We tackle the problem by requesting a small number of queries from each domain.

\section{Proposed method}\label{sec:prop}
\subsection{Semi-supervised gradual domain adaptation}\label{sec:prop-gda}
Since we consider the problem where the distances between the given domains are large, we propose a gradual domain adaptation method with a few labeled samples from the intermediate and the target domain. We can query $n_j$ samples from unlabeled pool dataset $U^{(j)}$. When the queried sample from the pool is annotated, each dataset in domain $j$ is updated as

\begin{equation}
\label{update_dataset}
D^{(j)} = D^{(j)} \cup (\bm{x}_i^{(j)}, y_i^{(j)}),\; U^{(j)} = U^{(j)} \backslash \bm{x}_{i'}^{(j)}.
\end{equation}
In this study, $n_j$ samples from each unlabeled dataset $U^{(j)}$ are given as initial values, and we add more labeled data by using multifidelity active learning. We describe the details of multifidelity active learning in Section~\ref{sec:prop-mfal}.
In our proposed method, we use only the labeled dataset $D^{(j+1)}$ from the next domain to train a model, and $\mathrm{SL}(D^{(j+1)})$ denotes the output of model training.
 
\begin{eqnarray}
    \mathrm{SL}(D^{(j+1)}) = \argmin_{\theta \in \Theta} \frac{1}{n_{j+1}} \sum_{i=1}^{n_{j+1}} \ell(h_{\theta}(\bm{x}_i^{(j+1)}), y_i^{(j+1)}).
    \label{eq:supervised}
\end{eqnarray}
When we train the model for the next domain, we use the parameter of the current domain $\theta^{j}$ as an initial value for the parameter of the next domain $\theta^{j+1}$. By utilizing a technique of warm-starting~\citep{erhan2010does}, we can train a model efficiently even when the number of labeled data is limited. We update the base model with the sequence of labeled datasets $D^{(1)}, D^{(2)}, \ldots, D^{(K)}$ and obtain the model parameters for all domains $\theta^{(1)}, \ldots, \theta^{(K)}$

\begin{align*}
    \theta^{(1)} &= \mathrm{SL}(D^{(1)}) \\
    \theta^{(2)} &= \mathrm{SL}(D^{(2)}) \\
    \vdots \\
    \theta^{(K)} &= \mathrm{SL}(D^{(K)}).
\end{align*}
We use the model parameters for each domain to calculate fidelities in multifidelity active learning.

\subsection{Multifidelity active learning}\label{sec:prop-mfal}
Since we consider the problem where the distances between the given domains are large, we propose a gradual domain adaptation method with a few labeled samples from the intermediate and the target domain. Although we can query from each domain, querying needs some cost $c^{(j)} \in \mathbb{R}, \; \bm{c} = (c^{(1)}, c^{(2)}, \ldots, c^{(K)})$. We consider the case that the reason why the target dataset has no label is because of its high cost. Moreover, it is natural to assume that the cost of obtaining data from the intermediate domain is higher when it is closer to the target domain and lower when it is closer to the source domain $c^{(1)} < c^{(2)} < \cdots < c^{(K)}$. Let $m^{(j)} \in \mathbb{N}, \; \bm{m} = (m^{(1)}, m^{(2)}, \ldots, m^{(K)})$ be the number of samples to query from the unlabeled dataset in the $j$-th domain. 

The cost $\bm{c}$ is given according to the problem, while the number of query samples $\bm{m}$ is determined from the dataset as explained later. Finally, there is a pre-fixed budget $B \in \mathbb{R}$, and the total cost cannot exceeds the budget, $\bm{m}\,\bm{c}^{T} \leq B$.

On the basis of the multifidelity Monte Carlo method~\citep{peherstorfer2016optimal, peherstorfer2018survey}, we considered the optimization of the number of query samples $\bm{m}$ in each domain by distributing budgets considering cost and fidelity. \citet{peherstorfer2016optimal} considered the optimization problem, which balances fidelity and cost such that the mean squared error of the multifidelity Monte Carlo estimator is minimized for a given budget. They introduce the ratio $\bm{r} = (r^{(1)}, \ldots, r^{(K)})$ of the number of query samples from the target domain to the number of query samples from the $j$-th domain. They showed that the ratio provides the global solution of the optimization problem. Note that $r^{(K)}$ is one since it is the ratio of the number of query samples from the target domain to itself. We show the definition of $r^{(s)}$ below:
\begin {align}
\label{r}
    r^{(s)} = \sqrt{\frac{c^{(K)}(\rho_s^{2} - \rho_{s-1}^{2})}{c^{(s)}(1 - \rho_{K-1}^{2})}}, \hspace{5mm} s = 1, \ldots, K-1,
\end {align}
where $\rho_s$ is the correlation coefficient between the output of the function $h_{\theta^{(K)}}(\bm{u})$ for the prediction of $\bm{x}^{(K)}$ and the output of the function $h_{\theta^{(j)}}(\bm{u})$ for the prediction of $\bm{x}^{(j)}$. Note that to evaluate $\rho_s$, we sampled the output of the function $h_{\theta^{(j)}}(\bm{u})$ by inputting randomly sampled $\bm{u} \in \mathbb{R}^{d}$. Each element of the sample $\bm{u}$ is uniformly chosen from the range of the observed maximum and minimum values of the original input data $\bm{x}$ in element-wise manner. When $s=1$, $\rho_{s-1} = 0$. 
From the budget and the calculated $\bm{r}$, the optimal number of queries $m^{(K)}$ from target domain is obtained by

\begin {align}
\label{mK}
    \tilde{m}^{(K)} = \frac{B}{\bm{r}\,\bm{c}^{T}}.
\end {align}
Using $\tilde{m}^{(K)}$, we calculate the number of queries from other domains by

\begin{align}
\label{m_other}
    \tilde{m}^{(s)} = r^{(s)} \tilde{m}^{(K)},\; s=1, \ldots, K-1.
\end{align}
We round down $\tilde{m}^{(1)}, \tilde{m}^{(2)}, \ldots, \tilde{m}^{(K)}$ to obtain an integer number and denote the optimal number of queries $\bm{m}$

\begin{align}
\label{m}
    \bm{m} = (m^{(1)}, m^{(2)}, \ldots, m^{(K)})=
    (\lfloor \tilde{m}^{(1)} \rfloor, \lfloor \tilde{m}^{(2)} \rfloor, \ldots, \lfloor \tilde{m}^{(K)} \rfloor).
\end{align}
We calculate the optimal number of queries whenever models are updated by adding new samples.

Finally, we add labels by active learning~\citep{DBLP:series/synthesis/2012Settles,DBLP:journals/corr/abs-2012-04225} for the number of samples obtained from each domain $m^{(j)}$. The samples to be queried are determined on the basis of uncertainty~\citep{ramirez2017active}, which is a standard measure in active learning. We compute the uncertainty of the samples in the unlabeled dataset $U^{(j)}$ of domain $j$ and query for the datum $x^{(j)\ast}$ with the largest uncertainty.

\begin {align}
    & {\mathrm{unc}}(\bm{x}_{i'}^{(j)}) = 1 - \displaystyle \max_{l \in \{ 1,\dots,L\}} \, \Pr(y=l | \bm{x}_{i'}^{(j)}, \theta^{(j)}), \label{unc} \\
    & \bm{x}^{(j)\ast} = \argmax_{\bm{x}_{i'}^{(j)} \in U^{(j)}} {\mathrm{unc}}(\bm{x}_{i'}^{(j)}). \label{query}
\end {align}
We describe the outline of our algorithm below, and the details of our algorithm are described in Algorithm~\ref{algo1}.
\begin{enumerate}
    \item We train the base model $h_{\theta^{(0)}}$ with the source dataset. 
    \item We update the base model with the sequence of labeled datasets $D^{(1)}, \ldots, D^{(K)}$ and obtain the model for all domains $h_{\theta^{(1)}}, \ldots, h_{\theta^{(K)}}$.
    \item Until the budget runs out, we query labels. We update all models $h_{\theta^{(1)}}, \ldots, h_{\theta^{(K)}}$ and recalculate the optimal number of queries $m^{(1)}, \ldots, m^{(K)}$ whenever one sample is queried from all domains. For an ablation study, we use random sampling instead of uncertainty sampling.
\end{enumerate}

\begin{algorithm}[!htbp]
\caption{Gradual Domain Adaptation with Multifidelity Active Learning}
\label{algo1}
\begin{algorithmic}[1]
\renewcommand{\algorithmicrequire}{\textbf{Input:}}
\renewcommand{\algorithmicensure}{\textbf{Output:}}
\algnewcommand\algorithmicforeach{\textbf{for each}}
\algdef{S}[FOR]{ForEach}[1]{\algorithmicforeach\ #1\ \algorithmicdo}
\Require datasets $D^{(0)}, \ldots , D^{(K)}$, and $U^{(1)}, \ldots , U^{(K)}$, budget $B$, cost $c^{(1)}, \ldots , c^{(K)}$
\Ensure models $h_{\theta^{(0)}}, \ldots, h_{\theta^{(K)}}$, updated datasets $D^{(1)}, \ldots, D^{(K)}$
\State compute the parameter of base model $\theta^{(0)}$ as Eq.~\eqref{eq:base-train}
\State initialize the number of queried samples from each domain $\bar{m}^{(1)}, \ldots, \bar{m}^{(K)} \leftarrow 0$
\While{$B > 0$}
\ForEach{$j \in \{0,\ldots, K\}$}
    \State train the model on the next domain $j+1$ with warm-starting
    \State minimize the loss given by Eq.~\ref{eq:supervised}
\EndFor
\State compute the optimal number of queries from each domain $m^{(1)}, \ldots, m^{(K)}$
\ForEach {$j \in \{1,\ldots, K\}$}
    \If {$\bar{m}^{(j)} < m^{(j)}$ and $B > c^{(j)}$}
    \State compute the uncertainty of the samples in the unlabeled dataset $U^{(j)}$ as Eq.~\eqref{unc}
    \State select the datum for query $x^{(j)\ast}$ by Eq.~\eqref{query}
    \State query the label of $x^{(j)\ast}$, and obtain $y^{(j)\ast}$
    \State update labeled and unlabeled dataset as Eq.~\eqref{update_dataset} 
    \State $D^{(j)} \leftarrow D^{(j)} \cup (\bm{x}_i^{(j)}, y_i^{(j)})$
    \State $U^{(j)} \leftarrow U^{(j)} \backslash \bm{x}_{i'}^{(j)}$
    \State $\bar{m}^{(j)} \leftarrow \bar{m}^{(j)} + 1$
    \State $B \leftarrow B - c^{(j)}$
    \EndIf
\EndFor
\EndWhile
\end{algorithmic}
\end{algorithm}

\section{Experiments}\label{sec:exp}
We evaluate the efficiency of the proposed method on four real-world datasets, and Pytorch~\citep{NEURIPS2019_9015} is used for all implementation. All experiments were conducted on Amazon Web Services (AWS)~\citep{AWS} Elastic Compute Cloud (EC2) instance whose type is g4dn.8xlarge. In the case of image dataset, the classifier $h_{\theta}$ was composed of three convolutional layers, one dropout, and one batch normalization layer. In the case of tabular dataset, the classifier $h_{\theta}$ was composed of two fully connected layers, one dropout, and one batch normalization layer. 
The results of twenty repetitions are shown for each experiment. Codes for reproducing the experiments are available at \url{https://github.com/ssgw320/gdamf}.

\subsection{Comparison between the previous and the proposed method}\label{sec:exp-previous}
\begin{figure}[!htbp]
\centering
  \includegraphics[clip, width=12cm]{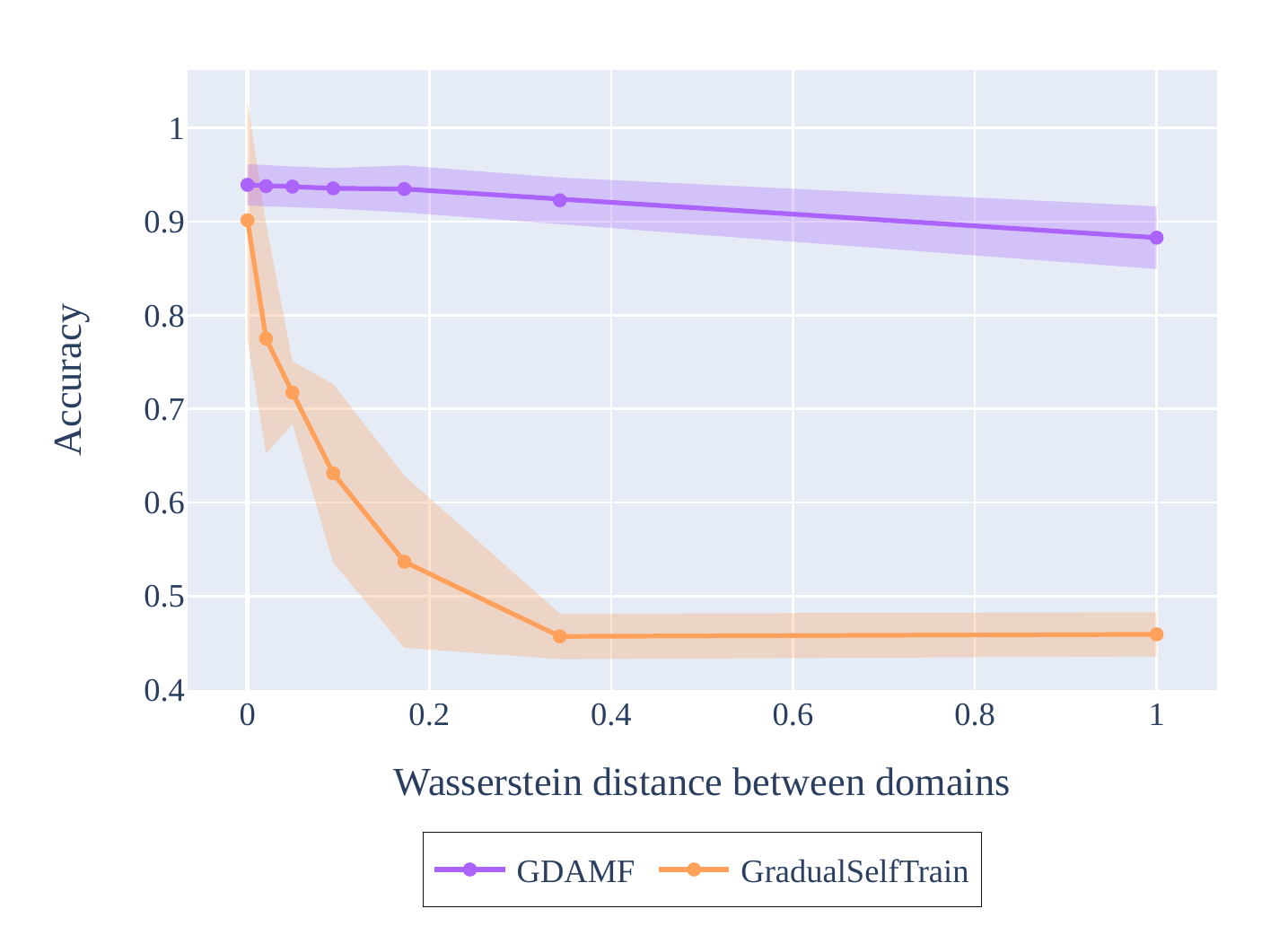}
  \caption{Comparison between the previous and the proposed method. The prediction performance of the previous method deteriorates when the distance of adjacent domains is large. In contrast, the proposed method achieves high accuracy even when the distance of adjacent domains is large.}\label{fig:two-moon}
\end{figure}

We compare our proposed method with the previous method~\citep{kumar2020understanding} on a toy dataset. The two-moon dataset is used as the source dataset, and the target dataset is prepared by rotating the source dataset by $\pi/2$. The size of source dataset $n_0$ is 2000. We adjust the distances of adjacent domains by changing the number of intermediate domains. For example, when the number of intermediate domains is two, the rotation angles of the intermediate domains are $\pi/6$ and $\pi/3$. We calculate a mean value of maximum per-class Wasserstein-infinity distance between adjacent domains when the number of intermediate domains is changed. We assign the minimum and the maximum number of the intermediate domain to 1 and 19, respectively. The results are shown in Figure~\ref{fig:two-moon}. Note that the distances shown in Figure~\ref{fig:two-moon} are scaled between zero and one. Although our proposed method needs queries, in this experiment, we set the budget as zero and run the algorithm only with initial samples from each domain since the two-moon dataset is easy to predict. We can consider various situations and settings of the number of initial samples. In this study, we assume that we cannot obtain a much number of initial samples and assign the number of initial samples to 1\% of the size of the source dataset $n_0/100$. The prediction performance of the previous method deteriorates when the distance of adjacent domains is large. In contrast, the proposed method achieves high accuracy by utilizing a small number of labeled samples even when the distance of adjacent domains is large. In Figure~\ref{fig:two-moon}, the distance of the maximum and the minimum number of the intermediate domains are plotted as 1 and 0, respectively. We can confirm that gradual domain adaptation with self-training is unsuitable when the distance between adjacent domains is lower than approximately 0.5.

\subsection{Datasets}\label{sec:exp-data}
We will modify existing datasets to emulate a gradual domain shift situation. Most of the datasets used in this paper are adopted from existing works~\citep{kumar2020understanding, abnar2021gradual}. We also add a gas sensor dataset~\citep{vergara2012chemical, rodriguez2014calibration}, which is considered more realistic gradual domain adaptation datasets.
\\
\noindent
Rotating MNIST: We prepared the rotating MNIST datasets, following the previous work~\citep{kumar2020understanding}. We prepared 2000 source datasets with no rotations, 42000 intermediate dataset with added gradually rotations of $\pi/63$ -- $\pi/3.15$, and 2000 target and evaluation datasets with added rotations of $\pi/3$. The intermediate dataset was divided into 20 parts.
\\
\noindent
Portraits: A dataset of actual photos of American high school students taken from 1905 to 2013 was used, and the task was to guess the gender from the images~\citep{ginosar2015century}. The dataset was sorted in descending order by year, with source, intermediate, target, and evaluation datasets beginning with the newest. All datasets had 2000 data each, and the intermediate had 6 domains in total.
\\
\noindent
Cover type: Cover type is a dataset available from the UCI repository. The task was to guess the type of plants in a group from 54 features~\citep{blackard1999comparative}. Following the previous work~\citep{kumar2020understanding}, we consider the binary classification problem by extracting only Spruce/Fir and Lodgepole Pine, which are the majority of the seven objective variables. In descending order of distance from the water body, the source, intermediate, target, and evaluation datasets were created. There are 8 intermediate domains in total, and each dataset had 50000 samples except the target and evaluation. The target and evaluation datasets had 30000 and 15000 samples, respectively.
\\
\noindent
Gas sensor: Gas sensor dataset is available from the UCI repository. The task was to guess the six types of chemicals from the responses of 128 sensors~\citep{vergara2012chemical, rodriguez2014calibration}. The data were divided into batches along the time the gas were processed, and shifts occurred as the sensor degraded along the batches. The last 3600 data were excluded because they were measured in very different environments. See~\citep{vergara2012chemical} for details. In ascending order of batch, the source, intermediate, target, and evaluation datasets were used. All datasets have 3000 data each except the evaluation, and we set the number of intermediate domain is one. The evaluation dataset had 1000 samples.

\begin{figure}[!htbp]
\centering
  \includegraphics[clip, width=12cm]{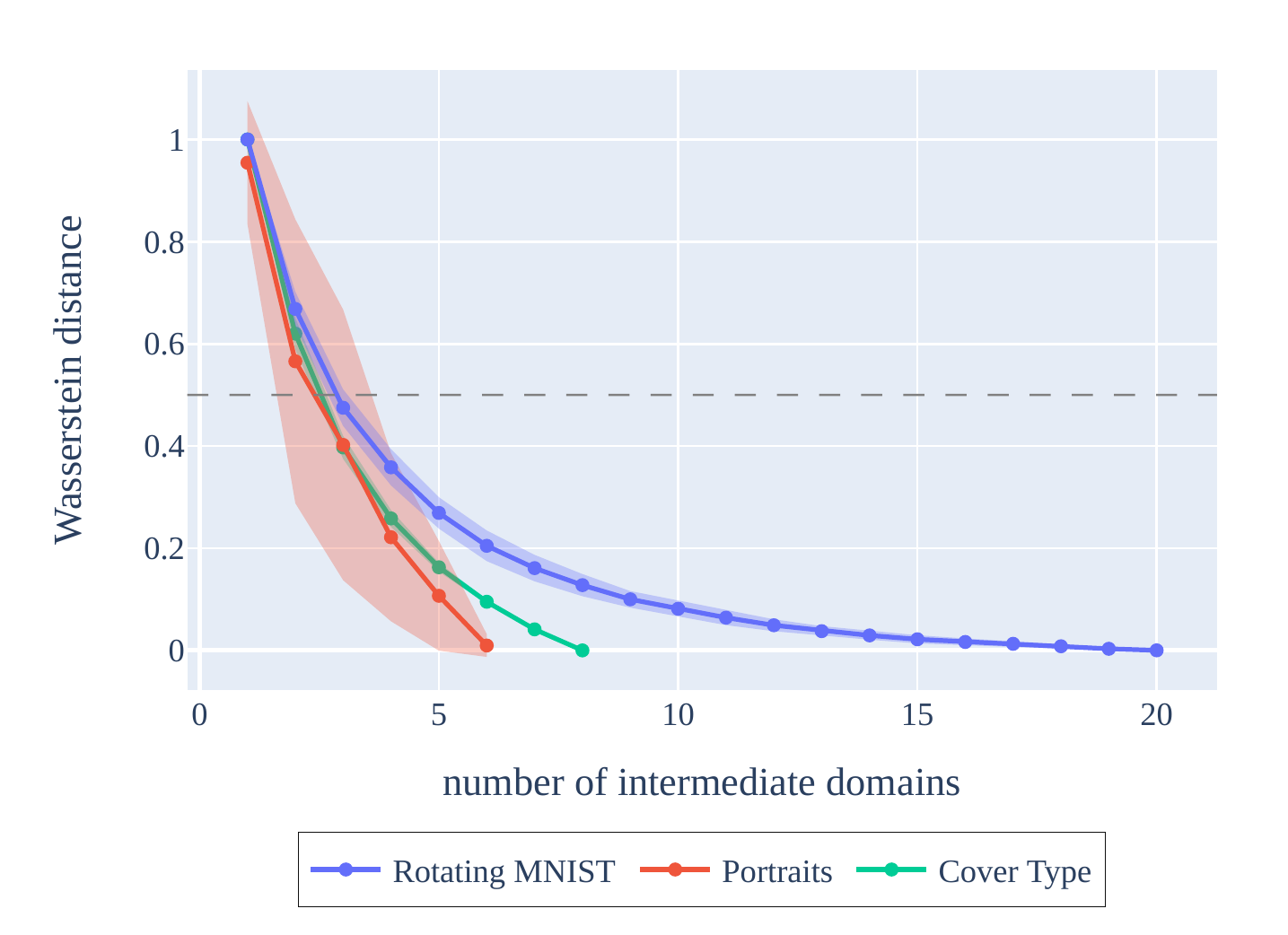}
  \caption{The distances between adjacent domains varies with the number of accessible intermediate domains. We randomly select an arbitrary number of intermediate domains from each dataset and calculate maximum per-class Wasserstein-infinity distances in adjacent domains.}\label{fig:wd}
\end{figure}

Since these datasets are used in the previous work~\citep{kumar2020understanding}, except the gas sensor, the distances between the domains are significantly small. Hence, we can apply gradual domain adaptation with self-training. On the other hand, we consider the problem where the distances between the adjacent domains are large. Thus, we restrict the number of accessible intermediate domains in each dataset. In practice, we cannot select arbitrary intermediate domains for domain adaptation. The sequence of intermediate domains is only given. In this study, we randomly select an arbitrary number of intermediate domains from each dataset. The arbitrary number is determined by the mean value of maximum per-class Wasserstein-infinity distances in adjacent domains. For example, when the number of accessible intermediate domains is five, we select five intermediate domains randomly and calculate the mean value of maximum per-class Wasserstein-infinity distances in adjacent domains. We show the calculation result with various numbers of accessible intermediate domains for each dataset in Figure~\ref{fig:wd}. Note that the distances are scaled between zero and one for each dataset. We select the number of accessible intermediate domains as two or three for each dataset since the number is closest to 0.5. We already confirmed in Section~\ref{sec:exp-previous} that the distance of 0.5 is unsuitable for self-training.

It is unnatural to assume that we can obtain the same number of labeled samples from the target domain as that of the source domain since we consider the case that the reason why the target dataset has no label is because of its high cost. In this study, we set 10\% of the source dataset size ${n_0}/{10}$ as the maximum budget. In the case of cover type, we set 1\% of the source dataset size ${n_0}/{100}$ as the maximum budget to conduct the experiment within a realistic time. We assign the indices of domains as cost. In the case of the rotating MNIST, the costs are $(c^{(1)}, c^{(2)}, c^{(3)}, c^{(4)}) = (1, 2, 3, 4)$ since we query from three intermediate datasets and one target dataset. Apart from the budget, we randomly select initial samples from each unlabeled dataset, and the number of initial samples is 1\% of the source dataset size ${n_0}/{100}$. Since the size of the source dataset is large, if we apply the same settings to cover type, the prediction performance is sufficiently high with initial samples only. Therefore, in cover type, we assign the number of initial samples to 0.1\% of the source dataset size. ${n_0}/{1000}$.

\subsection{Ablation study}\label{sec:exp-abl}
The features of our proposed method are to utilize intermediate domains and multifidelity active learning. We verify the effectiveness of our proposed method by ablation study. We describe the details of each ablation study below.\\
w/o AL: The query sample is determined at random.\\
w/o intermediate: The GDAMF algorithm runs only with the source and the target dataset.\\
w/o AL/intermediate: The GDAMF algorithm runs without intermediate datasets, and the query sample is determined randomly. We apply the warm-starting when the training of models.\\
w/o warm-starting: We do not apply the warm-starting when the training of models and assign a random value as the initial value of the model parameter.\\

\begin{figure}[!htbp]
\centering
  \includegraphics[clip, width=14cm]{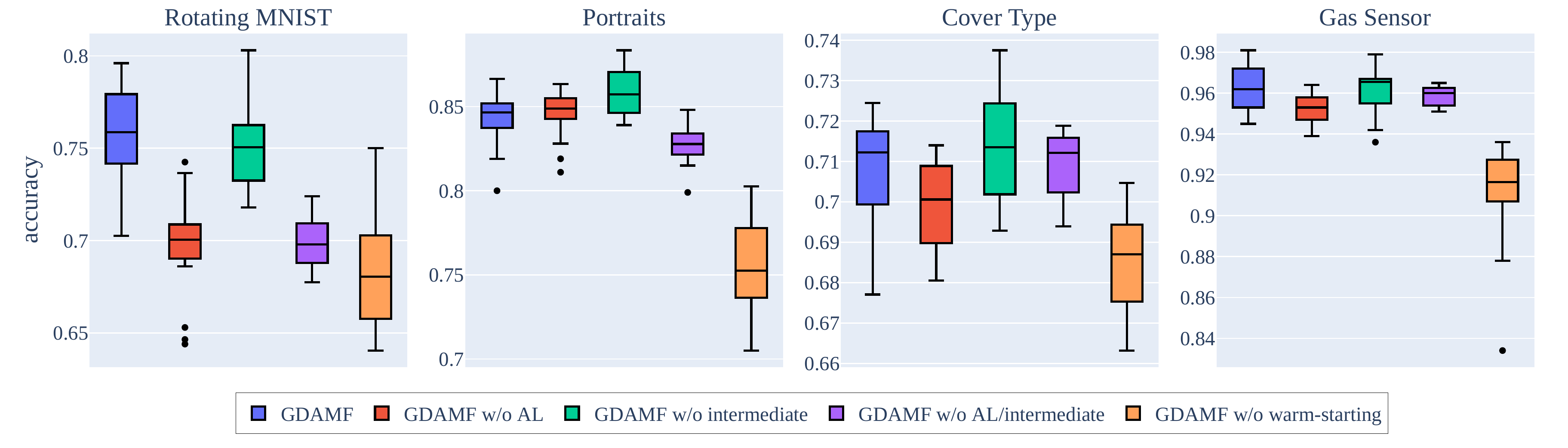}
  \caption{The result of the ablation study on real-world datasets. We confirm a deterioration of the prediction performance in both cases of w/o AL/intermediate and w/o warm-starting.}\label{fig:abl}
\end{figure}

We show the result of the ablation study in Figure~\ref{fig:abl}. We focus on the result of the rotating MNIST dataset and discuss the results of the ablation study since the rotating MNIST is a dataset whose gap between the source and the target domain is large. A significant deterioration of prediction performance is not confirmed in both cases of w/o AL and w/o intermediate. However, in both cases of w/o AL/intermediate and w/o warm-starting, we confirm a deterioration of the prediction performance. The contribution of warm-starting seems particularly large since the number of labeled samples from the intermediate and the target domains is small.

\begin{figure}[!htbp]
\centering
  \includegraphics[clip, width=14cm]{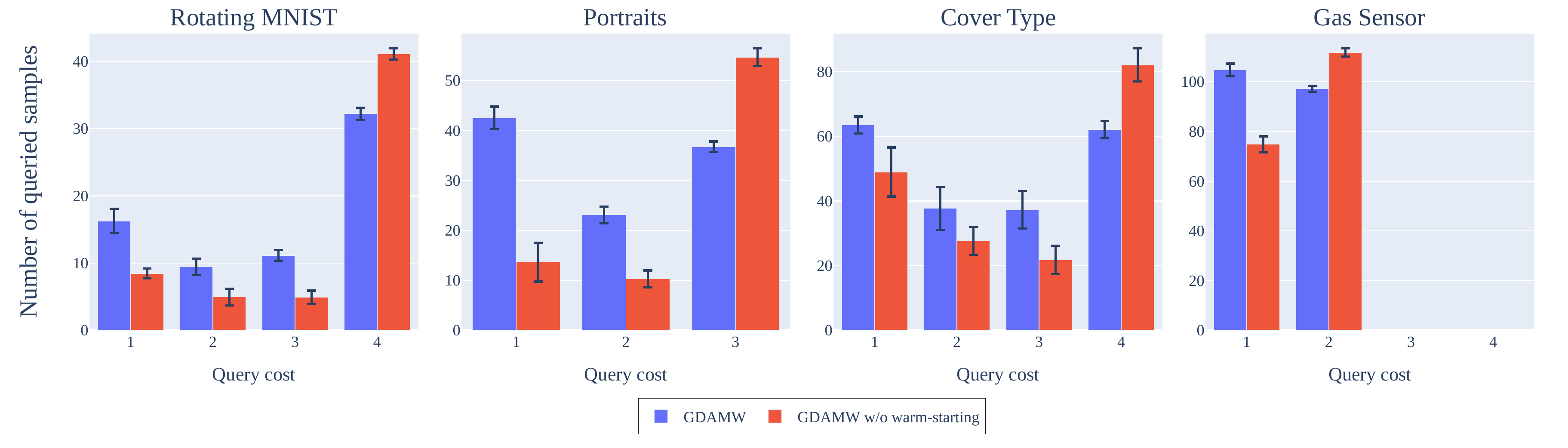}
  \caption{The number of queried samples from each domain with the maximum budget. When we run our algorithm without warm-starting, the number of queries from the target domain is large since the correlations between the models are small.}\label{fig:query}
\end{figure}

Figure~\ref{fig:query} shows the number of queried samples from each domain on the maximum budget. We can confirm that the multifidelity algorithm sets the number of query samples by considering the fidelity and cost in each domain. In the case of the portraits and the cover type dataset, the number of queries from the domains that are similar to the source domain is large since the gap between the source and the target is not large. When we run our algorithm without warm-starting, the number of queries from the target domain becomes large since the correlations between the models are weak. 

\subsection{Comparison with baseline method}\label{sec:exp-baseline}
Finally, we show that GDAMF can make more accurate predictions than existing baseline methods. DS-AODA~\citep{rai2010domain}, which is a previous work on active domain adaptation, is not a method for gradual domain adaptation, but it was used for comparison by updating the model step by step. We include GradualSelfTrain~\citep{kumar2020understanding}, GIFT~\citep{abnar2021gradual} and AuxSelfTrain~\citep{zhang2021gradual} in the comparative study because the implementations are publicly available. \citet{9681347} also proposed a gradual domain adaptation method. Although their proposed method requires four hyperparameters, it is not clear how those hyperparameters should be selected.

All the settings of experiments, such as model composition, query cost, and accessible intermediate domains, are the same as those of the proposed method. DS-AODA requires the probability of querying, whereas GIFT and AuxSelfTrain require the number of intermediate domains between the source and the target to be assumed as hyper parameters. Three levels, low, mid, and high, were evaluated for these hyperparameters. We abandon the evaluation of AuxSelfTrain algorithm with the cover type dataset since the algorithm needs more than 200GB of memory. Even if we have sufficient computer resources, the algorithm needs a huge amount of calculation, and the evaluation will not finish within a realistic time.
The experimental setup for each baseline method is shown below.\\
Target only: We randomly select data from the target dataset and train a model. The settings of budget and query cost are the same in the case of GDAMF.\\
DS-AODA: Update the model by querying the neighboring domains in order from source to target domains. The process ends when the budget is exhausted.\\  
GradualSelfTrain: Updates the model by self-training, giving intermediates in order from source to target domains. It does not depend on the budget.\\
GIFT: This method does not require intermediate domains; only source and target domains are given for training. It does not depend on the budget.\\
AuxSelfTrain: The intermediate and target domains are merged into a single target domain, and the source and target domains are given for training. It does not depend on the budget.\\ 

\begin{figure}[!htbp]
\centering
  \includegraphics[clip, width=14cm]{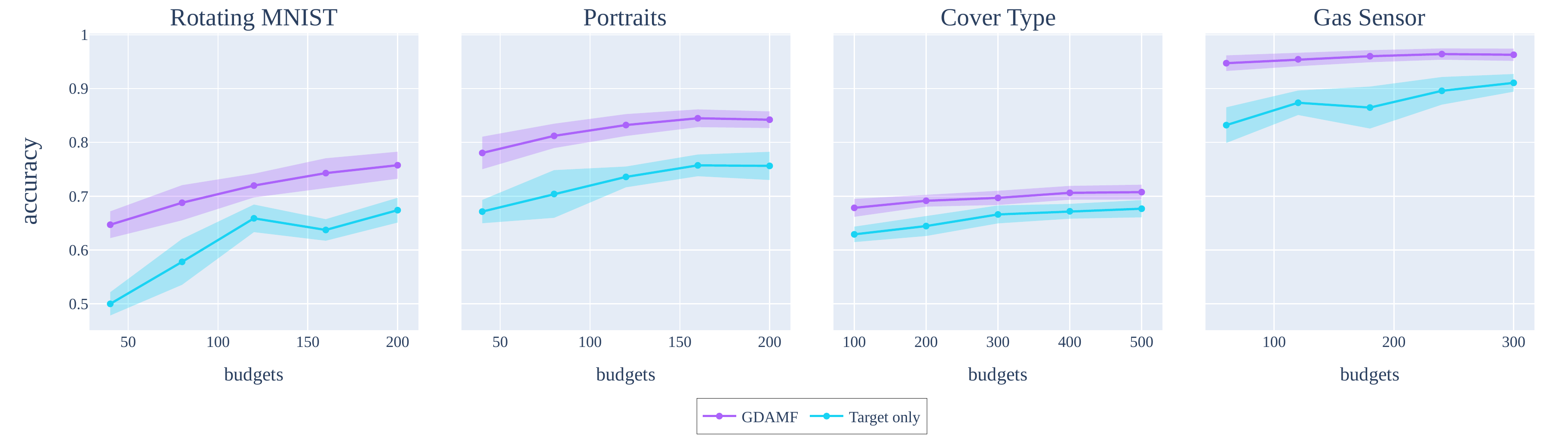}
  \caption{Experimental results with various budgets. The proposed method achieves better performance than ``Target only'' on all datasets, and we confirm that our proposed method is suitable for a problem with query cost.}\label{fig:budget}
\end{figure}

\begin{figure}[!htbp]
\centering
  \includegraphics[clip, width=14cm]{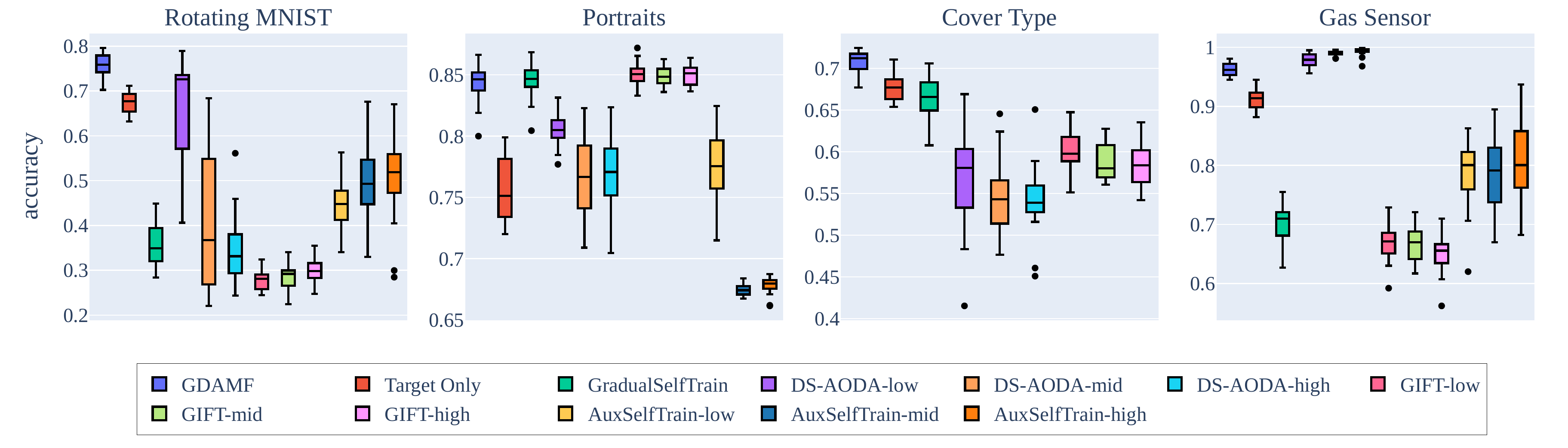}
  \caption{Comparison of classification accuracy on real-world datasets. Our proposed method is comparable to or superior to the baseline methods on all datasets.}\label{fig:baseline}
\end{figure}

We show the result with various budgets for each dataset in Figure~\ref{fig:budget}. The proposed method achieves better performance than ``Target only'' on all datasets, and we confirm that our proposed method is suitable for a problem with query cost.
Figure~\ref{fig:baseline} shows the results of twenty repetitions of training the model and calculating the accuracy on the test data. In the case of the method that requests a query, we show the results with the maximum budget. From the results in Figure~\ref{fig:baseline}, we see that the proposed method is comparable to or superior to the baseline methods on all datasets.

The proposed GDAMF stably achieves reasonable results and could be a strong candidate for the active gradual domain adaptation under multifidelity setting. In this study, we have run the DS-AODA algorithm sequentially for gradual domain adaptation. However, this approach has a possibility that the budget will run out until requesting a query from the target domain since this method does not control the number of queries from each domain. In the case of our proposed method, we can expect to avoid such a problem since our proposed method controls the number of query from each domain by utilizing multifidelity learning. Moreover, as a by-product, it can predict not only the target domain but also the intermediate domain.

\section{Discussion and Conclusion}\label{sec:disc&conc}
In gradual domain adaptation, the existence of an intermediate domain that gradually shifts from the source to the target domain is assumed. We consider the situation where the accessible intermediate domains are restricted, and propose GDAMF. By combining multifidelity and active domain adaptation, we addressed the trade-off between the query cost from each domain and the prediction accuracy using GDAMF. The effectiveness of the proposed method was evaluated on four real-world datasets, and it was shown that GDAMF provides reasonable performance under the condition that access to the intermediate domain is restricted and the query cost should be considered.

Although the proposed method provides high performance when there is a large domain gap between source and target, the effectiveness of the proposed method degrades when the domain gap between source and target is not large. 
In gradual domain adaptation, the ordered sequence of the intermediate domains is given, and it is assumed that there is no noisy intermediate domain whose correlations with other domains are low. Even if there is a noisy intermediate domain, our proposed method will distribute a budget to this domain. We should develop a method to select several appropriate intermediate domains from the initially given intermediate domains.
Since the multifidelity algorithm is only given information about fidelity and cost, even if there is a domain with low uncertainty, the algorithm will distribute a budget to this domain. We should set a stopping criterion for active learning~\citep{pmlr-v108-ishibashi20a} to mitigate this issue.

\section*{Acknowledgments}
Part of this work is supported by JST CREST Grant Nos. JPMJCR1761 and JPMJCR2015, JST-Mirai Program Grant No. JPMJMI19G1, and NEDO Grant No. JPNP18002.



\bibliographystyle{unsrtnat} 
\bibliography{ref}

\end{document}